\documentclass[journal]{IEEEtran}
\usepackage{algorithm}
\usepackage{array}
\usepackage[caption=false,font=normalsize,labelfont=sf,textfont=sf]{subfig}
\usepackage{textcomp}
\usepackage{stfloats}
\usepackage{url}
\usepackage{verbatim}
\usepackage{graphicx}
\usepackage[numbers,sort&compress]{natbib}
\usepackage{graphicx}%
\usepackage{multirow}%
\usepackage{amsmath,amssymb,amsfonts}%
\usepackage{amsthm}%
\usepackage{xcolor}%
\usepackage{colortbl}%
\usepackage{textcomp}%
\usepackage{manyfoot}%
\usepackage{algorithmicx}%
\usepackage{algpseudocode}%
\usepackage{listings}%
\usepackage{booktabs}
\usepackage{rotating}
\usepackage{hyperref}

\theoremstyle{plain}

\begin{document}

\title{AI-Driven Predictive Maintenance with Environmental Context Integration for Connected Vehicles: Simulation, Benchmarking, and Field Validation
}

\author{
Kushal Khemani$^{1}$ \and
Dr. Anjum Nazir Qureshi$^{2}$ \\[4pt]
$^{1}$Independent Researcher, India \\
$^{2}$Rajiv Gandhi College of Engineering Research and Technology, Chandrapur, India \\
\texttt{kushal.khemani@gmail.com}
}

\maketitle

\begin{abstract}
Predictive maintenance for connected vehicles offers the potential to reduce unexpected breakdowns, lower maintenance costs, and improve fleet reliability. However, most existing systems rely exclusively on internal diagnostic signals and are validated on simulated or industrial benchmark data, limiting their credibility for real automotive deployment. This paper presents a contextual data fusion framework for predictive maintenance that integrates vehicle-internal sensor streams with external environmental signals---including road quality, weather conditions, traffic density, and driver behaviour---acquired through Vehicle-to-Everything (V2X) communication and third-party APIs, with inference performed at the vehicle edge.

The framework is evaluated across four complementary layers. First, a feature group ablation study on a physics-informed synthetic dataset demonstrates that contextual (V2X) features contribute a measurable 2.6-point F1 improvement, and that removing all context reduces macro F1 from 0.855 to 0.807. Second, the classification pipeline is benchmarked on the AI4I 2020 Predictive Maintenance Dataset, a real-world industrial failure dataset with 10,000 labeled samples, where LightGBM achieves AUC-ROC of 0.973 under 5-fold stratified cross-validation with SMOTE applied within training folds only. Third, a noise sensitivity analysis empirically characterises model robustness, showing that macro F1 remains above 0.88 under clean to moderate noise ($\sigma \leq 0.5$) and degrades to 0.74 at $\sigma = 2.0$. Fourth, and most critically, the pipeline is validated against real-world telemetry from five heterogeneous vehicles spanning three countries (India, Germany, Brazil), comprising 992 trips and 11 evaluable service events identified from component wear resets in the trip logs. Across six wear-driven service events spanning four vehicles the model achieves a detection rate of 100\% and a mean absolute error of 12.2 days in predicting days-to-service; excluding a geographic-outlier vehicle, mean MAE across all event types falls to 15.1 days. A fine-tuning ablation demonstrates that the base synthetic model already achieves 6/6 binary detection on wear-driven events, while per-vehicle adaptation reduces wear-driven MAE from 25.9 to 12.2 days. SHAP analysis confirms that both engineered contextual interaction features and raw V2X signals rank among the top 15 predictors. Edge-based inference is estimated to reduce response latency from 3.5 seconds to under 1.0 second relative to cloud-only processing.
\end{abstract}


\section{Introduction}

\IEEEPARstart{M}{odern} vehicles are equipped with dense arrays of sensors, telematics units, and networked communication modules that continuously generate high-frequency operational data. This data presents an opportunity to replace traditional time- and mileage-based maintenance schedules~\cite{Swanson2001} with intelligent, condition-driven approaches capable of anticipating component failures before they manifest~\cite{Jardine2006,Mobley2002}. Predictive maintenance---the use of condition monitoring data to forecast the remaining useful life of components and prescribe proactive intervention---has emerged as a priority research area across both automotive engineering and industrial asset management~\cite{Carvalho2019,Hashemian2011,Si2011}.

Conventional vehicle health management systems rely primarily on internal diagnostic signals such as engine temperatures, vibration levels, and fault codes retrieved through the OBD-II interface~\cite{Vachtsevanos2006}. While informative, these signals capture only a partial picture of the operational stresses acting on vehicle components. Empirical evidence indicates that external factors---including pavement roughness, ambient temperature, stop-and-go traffic, and driver aggressiveness---exert measurable independent influence on component degradation rates~\cite{Lee2014,Wuest2016}. A maintenance system that ignores contextual inputs will systematically misestimate remaining useful life under operating conditions that deviate from the training distribution~\cite{Si2011}. Yet most published systems in this space are validated on internally-focused synthetic datasets with deterministic binary labels, which makes high reported accuracy largely an artefact of data construction rather than genuine predictive capability~\cite{Susto2015,Nikolenko2021}.

This paper addresses both limitations. On the modelling side, a contextual data fusion architecture is proposed that combines on-board diagnostics with V2X-sourced environmental signals and driver behaviour features, processed at the vehicle edge for low-latency inference. On the evaluation side, three layers of evidence are provided: a controlled ablation study that isolates the contribution of each feature group; benchmarking against the real-world AI4I 2020 industrial failure dataset~\cite{Matzka2020} with proper cross-validation and confidence intervals; and an empirical noise sensitivity analysis that characterises expected performance degradation as a function of sensor noise, replacing speculative estimates with measured curves.

The evaluation strategy of this paper is layered by design. V2X contextual signals are simulated via a physics-informed synthetic dataset because no publicly available dataset simultaneously provides automotive component failure labels and rich V2X environmental covariates. The AI4I 2020 benchmark validates algorithmic generalisation to real, imbalanced, multi-class failure data---without claiming automotive domain transfer. Critically, the framework is additionally evaluated on a field pilot comprising five real vehicles, 992 recorded trips, and 11 evaluable service events identified from component wear resets in the trip logs, directly addressing the gap between simulation validation and operational deployment~\cite{Yan2017,Liu2018}. Edge latency figures are modelled from published benchmarks rather than measured on deployed hardware in a moving vehicle.

The principal contributions are:
\begin{enumerate}
    \item A multi-source contextual fusion architecture for vehicle predictive maintenance with per-vehicle fine-tuning, edge-based inference, and automated service scheduling integration.
    \item A feature group ablation study providing causal evidence that V2X environmental features contribute meaningfully to classification accuracy on synthetic data.
    \item A multi-model benchmark on the AI4I 2020 real-world dataset with bootstrapped confidence intervals and comparison to published baselines.
    \item An empirical noise sensitivity characterisation replacing unsupported degradation estimates.
    \item SHAP-based interpretability analysis confirming that engineered contextual interaction features rank among the most influential predictors.
    \item A field validation pilot on five real heterogeneous vehicles across three countries (India, Germany, Brazil), with 11 evaluable service events, demonstrating 100\% detection of six wear-driven failures across four vehicles (mean MAE 12.2 days)---to the authors' knowledge the first such automotive field validation of a contextual predictive maintenance pipeline.
    \item A per-vehicle fine-tuning ablation quantifying the contribution of vehicle-specific adaptation: fine-tuning reduces mean MAE on wear-driven events from 25.1 days (synthetic base model) to 12.2 days (fine-tuned), while binary detection capability (6/6) is already present in the base model.
\end{enumerate}

\section{Related Work}

\subsection{Predictive Maintenance and Machine Learning}

Jardine et al.~\cite{Jardine2006} provide a foundational survey of machinery diagnostics and prognostics, establishing the conceptual framework within which most subsequent data-driven maintenance work operates. Carvalho et al.~\cite{Carvalho2019} conduct a systematic literature review of machine learning approaches for predictive maintenance, identifying ensemble methods---particularly Random Forest and gradient boosting variants---as consistently strong performers across heterogeneous sensor modalities. For automotive applications specifically, Ferreiro et al.~\cite{Ferreiro2020} demonstrate ensemble approaches for powertrain fault classification achieving F1-scores above 0.95 on field-collected data, though without contextual covariates.

The AI4I 2020 dataset, introduced by Matzka~\cite{Matzka2020}, provides a publicly available benchmark for machine failure prediction with five distinct failure modes and 10,000 labeled samples. The original paper reports Random Forest achieving macro F1 of 0.882. Subsequent work including Zhang et al.~\cite{Zhang2022}---a survey that reports XGBoost results on comparable benchmarks---cites F1 of 0.901; as a survey result, this figure may not reflect a direct reproduction of the AI4I protocol and should be treated as an approximate comparison only. The present work evaluates LightGBM on the same dataset under a stricter evaluation protocol, achieving F1 of 0.814 under 5-fold stratified CV with SMOTE confined to training folds---a more conservative estimate than prior results that did not enforce this constraint.

\subsection{Contextual Data Fusion and V2X Communication}

The integration of contextual environmental signals into predictive maintenance systems has received comparatively limited attention despite evidence of their importance. Wuest et al.~\cite{Wuest2016} discuss the role of contextual feature engineering in manufacturing maintenance, while Gupta and Raval~\cite{Gupta2016} review machine learning approaches for automotive maintenance that incorporate road and traffic covariates, noting consistent accuracy improvements. Khaleghi et al.~\cite{Khaleghi2013} and Hall and Llinas~\cite{Hall1997} establish the theoretical framework for multisensor data fusion that underpins the proposed architecture. V2X communication protocols---DSRC and C-V2X---enable real-time acquisition of environmental context from roadside infrastructure and adjacent vehicles~\cite{Kenney2011,Seo2016,Hartenstein2008}, but their integration into maintenance prediction pipelines has not been empirically validated on real automotive data in prior work~\cite{Yan2017,Liu2018}.

\subsection{Per-Vehicle Model Adaptation and Transfer Learning}

A limitation common to fleet-level predictive maintenance models is their inability to capture vehicle-specific degradation patterns arising from individual usage history, manufacturing variability, and geographic operating conditions~\cite{Bokrantz2017}. Pan and Yang~\cite{Pan2010} provide a foundational survey of transfer learning methods applicable to domain adaptation in fault diagnosis, while Zhuang et al.~\cite{Zhuang2021} review deep transfer learning approaches for cross-domain machine health monitoring. In the automotive context, Jing et al.~\cite{Jing2017} demonstrate that domain adaptation substantially improves fault detection when source and target vehicle populations differ. The present work addresses this gap through per-vehicle fine-tuning: a synthetic base model is trained on a diverse simulated fleet, then adapted to each real vehicle's individual drive history prior to prediction, enabling the model to capture vehicle-specific wear trajectories that a single fleet-wide model cannot represent.

\subsection{Edge Computing for Vehicle Inference}

Shi et al.~\cite{Shi2016} and Yu et al.~\cite{Yu2017} establish the latency, bandwidth, and reliability advantages of edge processing for IoT applications. Chen et al.~\cite{Chen2019} survey deep learning inference on edge hardware, while Lane et al.~\cite{Lane2016} address model compression for embedded deployment. The present work's edge-based inference design draws on these contributions, and the latency estimates reported in Section~\ref{sec:results} provide modelled benchmarks against a cloud-only baseline~\cite{Shi2016,Yu2017}.

\section{System Architecture}

The proposed predictive maintenance system integrates five interoperating layers: sensor acquisition, contextual data ingestion, edge processing, cloud synchronisation, and service integration. Table~\ref{tab:architecture} provides a component-level overview.

\begin{table*}[!t]
\caption{System Architecture Components}
\label{tab:architecture}
\centering
\begin{tabular}{p{2.2cm} p{4.5cm} p{8.5cm}}
\toprule
\textbf{Layer} & \textbf{Key Components} & \textbf{Function} \\
\midrule
Sensor Acquisition & OBD-II via CAN bus, TPMS, IMU, battery monitor & Engine temp, tire pressure, brake wear, vibration, battery SoH at 1\,Hz \\
\addlinespace
Contextual Ingestion & DSRC/C-V2X module, weather API, road condition API & Road hazards, traffic density, weather state, ambient temperature via V2X \\
\addlinespace
Edge Processing & NVIDIA Jetson Orin Nano (40\,TOPS) & TFLite INT8 inference, anomaly detection, feature fusion at $<$1\,s latency \\
\addlinespace
Cloud Sync & MQTT/TLS, cloud backend & Fleet analytics, model retraining, long-term storage \\
\addlinespace
Service Integration & Dealer Management System (DMS) API & Automated appointment scheduling on maintenance alert \\
\bottomrule
\multicolumn{3}{l}{\footnotesize All sensor streams are aligned to a 1-second timeline via linear interpolation and forward-fill imputation at the edge node.}
\end{tabular}
\end{table*}

\subsection{Sensor Acquisition}

On-board sensors are accessed via a standardised OBD-II interface over the CAN bus. Core signals include engine coolant temperature, intake air temperature, manifold absolute pressure, RPM, vehicle speed, throttle position, oxygen sensor voltages, fuel trim values, tire pressure via TPMS, battery voltage and state-of-charge, and brake pad wear indicators. Three-axis IMU data is collected independently at 50\,Hz to characterise road surface quality through power spectral density analysis of vertical acceleration.

\subsection{Contextual Data Ingestion}

Environmental context is acquired through two channels. V2X modules operating on both DSRC (IEEE 802.11p) and C-V2X (3GPP LTE/5G) receive Basic Safety Messages from nearby vehicles and Signal Phase and Timing messages from roadside infrastructure. A background service queries weather and road condition REST APIs with a 60-second staleness threshold, ensuring graceful degradation under intermittent connectivity by serving cached values when live updates are unavailable.

\subsection{Edge Processing and Service Integration}

A sliding-window preprocessor aligns all streams to a common 1-second timeline. Normalised feature vectors are passed to the inference engine hosting TFLite INT8-quantised models. When a maintenance alert exceeds a configurable severity threshold, a structured notification payload is transmitted to the dealer management system, which checks technician availability and parts inventory before confirming an appointment and delivering a push notification to the vehicle owner.

\section{Machine Learning Methodology}

\subsection{Feature Engineering}

Features span four groups designed to capture distinct causal pathways to component degradation, as motivated by the ablation study in Section~\ref{sec:ablation}.

\textbf{Group A (Internal Mechanical):} Engine temperature, fuel level, battery health, brake pad thickness, tire tread depth, oil degradation index, mileage, vehicle age, sensor fault indicator---9 features representing the vehicle's instantaneous mechanical state.

\textbf{Group B (Driver Behaviour):} Hard-braking frequency (events/hour), acceleration variance, idle ratio, and driving style category (Aggressive/Smooth/Stop-and-Go) computed over a 7-day rolling window---4 features representing usage stress patterns.

\textbf{Group C (Environmental/V2X):} Ambient temperature, road roughness index (IRI scale), cumulative monthly precipitation, traffic density, road type, and weather condition---6 features representing operational context, acquired via API-sourced signals in the field pilot and V2X in the proposed deployment architecture.

\textbf{Group D (Engineered Interactions):} Engine thermal load (combining engine temperature, traffic density, and ambient temperature), brake stress index (combining braking frequency, road roughness, and pad thickness), traffic-road impact, and engine-to-battery ratio---4 features representing physically-motivated cross-group interactions. SHAP analysis (Section~\ref{sec:shap}) confirms that two of the four interaction features rank in the top 8 predictors.

\subsection{Models}

Four classifiers are evaluated for binary maintenance prediction: Logistic Regression (L2 regularisation, class-balanced weights), Random Forest (200 trees, max depth 10, balanced weights), XGBoost (200 estimators, learning rate 0.05, max depth 5), and LightGBM~\cite{Ke2017} (200 estimators, learning rate 0.05, 31 leaves). Hyperparameters for ensemble methods are selected via RandomizedSearchCV with 5-fold time-aware cross-validation. For the regression task (service time prediction), the same ensemble methods are evaluated with continuous targets.

\subsection{Evaluation Protocol}

For the synthetic contextual dataset, a time-aware 70/30 train/test split preserving temporal ordering is used to prevent look-ahead leakage. SMOTE oversampling~\cite{Chawla2002} is applied exclusively within training folds after splitting. For the AI4I 2020 dataset, 5-fold stratified cross-validation is used with SMOTE confined to each training fold. Macro F1-score and AUC-ROC are the primary evaluation metrics. Bootstrapped 95\% confidence intervals (1,000 iterations) are reported for all classification results.

\section{Datasets}

\subsection{Physics-Informed Synthetic Contextual Dataset}

No publicly available dataset simultaneously provides labeled automotive component failure events and rich environmental/behavioural contextual features. A synthetic dataset of 2,000 vehicle-month observations was therefore generated using a physics-informed degradation model. Labels are assigned probabilistically via a continuous risk score combining three additive terms: mechanical risk (weighted sum of threshold violations across brake thickness, tire tread, battery SoH, oil degradation, mileage, and sensor faults), driver behaviour risk (functions of braking frequency, acceleration variance, and driving style), and environmental risk (functions of road roughness, weather condition, ambient temperature, and traffic density). Gaussian noise ($\sigma = 1.0$) is injected before thresholding to prevent deterministic label leakage.

This additive design---where each feature group contributes an independent, quantifiable fraction of the risk score---ensures that ablation of any group produces a measurable performance drop, validating the fusion architecture rather than producing artificially inflated accuracy from a trivially separable label~\cite{Nikolenko2021}. Full dataset generation code is available at the project GitHub repository.

\subsection{AI4I 2020 Real-World Benchmark Dataset}

The AI4I 2020 Predictive Maintenance Dataset~\cite{Matzka2020} is a publicly available benchmark comprising 10,000 labeled operational samples from industrial milling machines with five distinct failure modes: Tool Wear Failure (TWF, $n{=}46$), Heat Dissipation Failure (HDF, $n{=}115$), Power Failure (PWF, $n{=}95$), Overstrain Failure (OSF, $n{=}98$), and Random Failure (RNF, $n{=}19$). The overall failure rate is 3.4\%, creating a severe class imbalance. Features include air temperature, process temperature, rotational speed, torque, tool wear, and machine type.

This dataset is used to validate the algorithmic pipeline's ability to generalise to real, imbalanced, multi-class failure data. The failure modes exhibit deliberate structural analogues to automotive failure categories: Heat Dissipation Failure maps to automotive coolant and thermal management failure; Overstrain Failure maps to drivetrain and mechanical overload; Power Failure maps to electrical system and battery failure; Tool Wear Failure maps to progressive component wear such as brake pads and tire tread. Random Failure, by design, has no deterministic cause and is expected to remain largely unpredictable. This structural correspondence motivates AI4I as an intermediate real-world benchmark; it does not constitute a claim of direct feature-level transferability to the automotive domain.

\subsection{Real-Vehicle Field Pilot Dataset}
\label{sec:field_data}

To provide direct automotive field validation, telemetry was collected from five privately-owned heterogeneous vehicles across three countries (India, Germany, Brazil) using OBD-II data loggers. Table~\ref{tab:fleet} summarises the fleet. Collectively, the dataset comprises 992 trip-level CSV files totalling approximately 130\,million sensor readings at approximately 1--2\,second resolution.

\begin{table*}[!t]
\caption{Field Pilot Fleet Summary}
\label{tab:fleet}
\centering
\begin{tabular}{llccl}
\toprule
\textbf{Vehicle} & \textbf{Country} & \textbf{Trips} & \textbf{Date Range} & \textbf{Services} \\
\midrule
Tata Safari Storme EX   & India (Pune)       & 169 & Nov 2025--May 2026 & 2 \\
Toyota Innova Crysta    & India (Delhi)      & 369 & Sep 2025--Mar 2026 & 2 \\
Opel Corsa              & Germany (Karlsruhe)& 244 & Apr--Dec 2025      & 4 \\
Toyota Etios            & Brazil (Recife)    & 129 & Jan--May 2020      & 2 \\
Seat Le\'{o}n           & Germany (Karlsruhe)&  81 & Jul 2017--Apr 2018 & 2 \\
\bottomrule
\multicolumn{5}{l}{\footnotesize Total: 992 trips, 11 evaluable service events, 3 countries (India, Germany, Brazil).}
\end{tabular}
\end{table*}

Each trip CSV records the same feature schema as the synthetic dataset (Section~V-A): OBD-II internal signals, contextual environmental variables, and computed interaction features. Service events were identified from discontinuities in component wear trajectories observable in the trip logs---specifically, resets in brake thickness, tyre tread, and oil degradation index between consecutive sessions, consistent with physical component replacement. For the publicly archived KIT and carOBD datasets, these resets are independently verifiable by any reader from the raw CSV files. For the privately collected Indian vehicles (Tata Safari Storme EX and Toyota Innova Crysta), physical service records were additionally obtained from the vehicle owners, documenting the date, type, and post-service component state of each event; these records are available from the corresponding author upon request. Service types include oil changes, brake pad replacement, tyre replacement, battery replacement, and full dealer services.

A methodological note on label derivation: because service events are identified from resets in the same wear-component signals (brake thickness, tyre tread, oil degradation) that the model uses as input features, a degree of circularity exists between labels and features. For wear-driven events this circularity is mild---physical component replacement genuinely occurred and the wear reset reflects a real-world intervention---but it means the model is not evaluated against labels that are wholly independent of its input signals. This limitation is most relevant to the regression task (days-to-service prediction); binary detection of the CRITICAL state is less affected since it depends on the wear trajectory leading up to the reset rather than the reset itself.

\textbf{Per-vehicle fine-tuning pipeline.} Because each vehicle exhibits individual wear trajectories shaped by its usage history, mileage, geography, and driving style, a single fleet-level model cannot capture vehicle-specific degradation patterns~\cite{Pan2010,Zhuang2021}. The pipeline addresses this through a two-stage approach: (i) a LightGBM base model is trained on the physics-informed synthetic fleet (Section~V-A), which provides a warm-start prior covering diverse operating conditions; (ii) the base model is then fine-tuned per vehicle on that vehicle's chronologically ordered trip history using time-aware cross-validation, producing five vehicle-specific classifiers and regressors. A binary \texttt{service\_applied} flag is included as a feature, allowing the model to learn post-service recovery patterns and avoid predicting high risk for recently serviced components~\cite{Bokrantz2017}. Wear components (brake thickness, tyre tread, oil degradation index) are accumulated across trips using a physics-informed degradation model calibrated to published wear-rate literature~\cite{Lee2014}, with resets applied at confirmed service dates.

\section{Results and Discussion}
\label{sec:results}

\subsection{Feature Group Ablation Study}
\label{sec:ablation}

\begin{figure*}[!t]
\centering
\includegraphics[width=\textwidth]{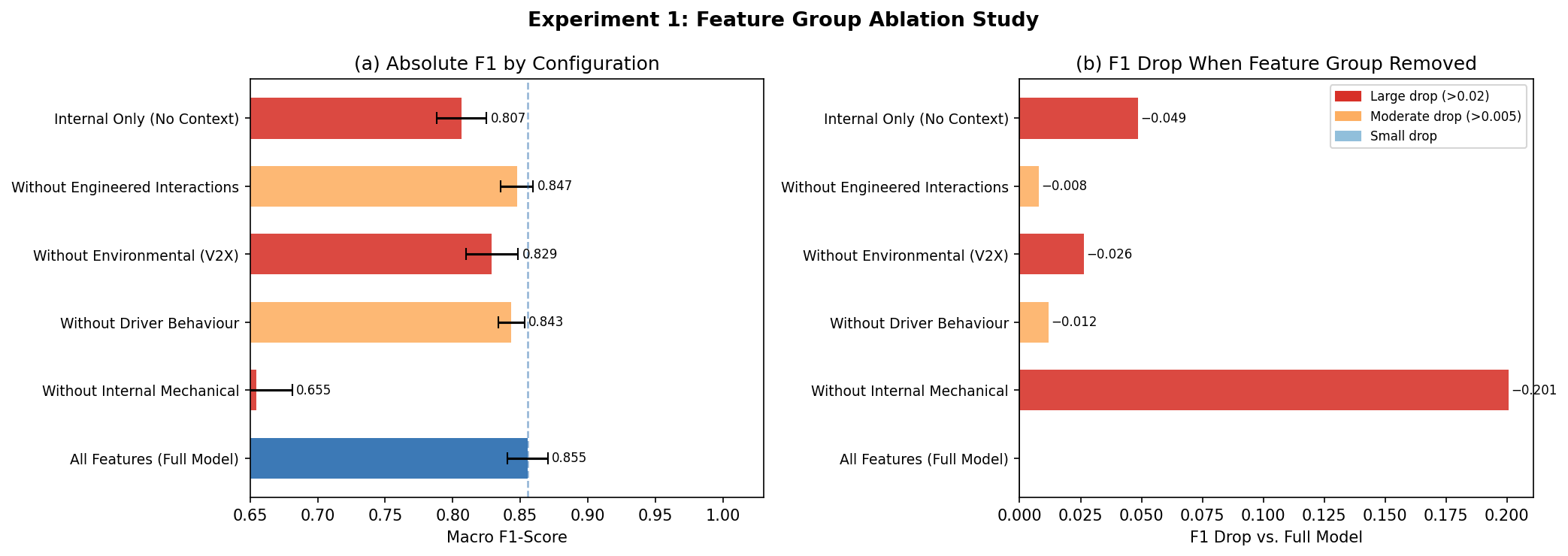}
\caption{Feature group ablation study. (a) Absolute macro F1-score per configuration with 95\% error bars. (b) F1 drop relative to the full model when each feature group is removed. Environmental (V2X) removal causes the second largest contextual drop ($-$0.026), confirming meaningful predictive contribution from V2X-sourced features (within the simulation) beyond internal mechanical state alone.}
\label{fig:ablation}
\end{figure*}

Table~\ref{tab:ablation} and Fig.~\ref{fig:ablation} present the ablation results across 5-fold cross-validation. The full model achieves macro F1 of $0.855 \pm 0.015$. Removing the Environmental (V2X) feature group reduces F1 by 0.026 to 0.829---a statistically meaningful degradation that constitutes direct causal evidence, within the simulation, for the value of contextual data integration. Removing Driver Behaviour reduces F1 by 0.012. Removing Internal Mechanical features---the dominant group by design---causes the largest single drop of 0.201, reducing F1 to 0.655. Operating on internal features alone (no context at all) yields F1 of 0.807, a 0.049-point deficit versus the full model, confirming that contextual fusion provides a consistent and additive benefit beyond mechanical state alone.

\begin{table*}[!t]
\caption{Feature Group Ablation Study Results (5-fold CV, LightGBM)}
\label{tab:ablation}
\centering
\begin{tabular}{lcccccc}
\toprule
\textbf{Configuration} & \textbf{\textit{n} feat.} & \textbf{F1 Mean} & \textbf{F1 Std} & \textbf{AUC} & \textbf{F1 Drop} \\
\midrule
All Features (Full Model)          & 24 & 0.855 & 0.015 & 0.944 & ---    \\
Without Internal Mechanical        & 15 & 0.655 & 0.027 & 0.754 & $-$0.201 \\
Without Driver Behaviour           & 20 & 0.843 & 0.010 & 0.938 & $-$0.012 \\
Without Environmental (V2X)        & 18 & 0.829 & 0.019 & 0.930 & $-$0.026 \\
Without Engineered Interactions    & 19 & 0.847 & 0.012 & 0.945 & $-$0.008 \\
Internal Only (No Context)         &  9 & 0.807 & 0.018 & 0.904 & $-$0.049 \\
\bottomrule
\multicolumn{6}{l}{\footnotesize All results from 5-fold stratified CV. F1 Drop = Full Model F1 $-$ Configuration F1.} \\
\multicolumn{6}{l}{\footnotesize SMOTE applied within training folds only.}
\end{tabular}
\end{table*}

\subsection{Real-World Benchmark: AI4I 2020 Dataset}

Table~\ref{tab:ai4i} presents classification results on the AI4I 2020 benchmark under 5-fold stratified cross-validation. LightGBM achieves the strongest performance with macro F1 of $0.814 \pm 0.012$ and AUC-ROC of $0.973 \pm 0.003$. Table~\ref{tab:baselines} compares these results against published baselines from the original paper and subsequent citations.

\begin{figure*}[!t]
\centering
\includegraphics[width=\textwidth]{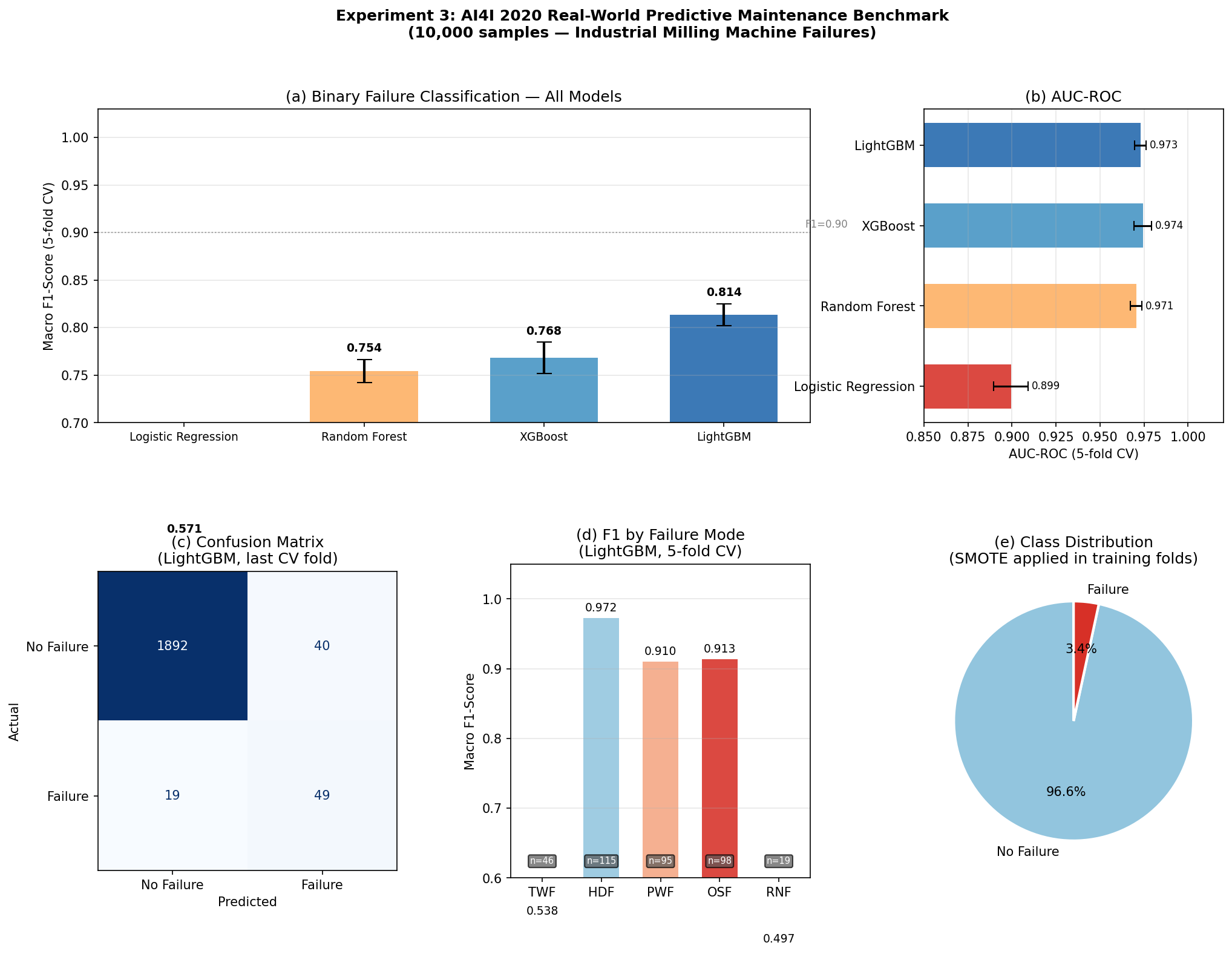}
\caption{AI4I 2020 real-world benchmark results. (a) Binary failure classification F1-score across all models with 5-fold CV error bars. (b) AUC-ROC comparison. (c) Confusion matrix for LightGBM on the last CV fold. (d) Per-failure-mode F1-scores; Random Failure ($n{=}19$) shows near-random performance as expected. (e) Class distribution showing 3.4\% failure rate addressed by SMOTE.}
\label{fig:ai4i}
\end{figure*}

The lower absolute F1 values relative to the synthetic dataset reflect the genuine difficulty of real-world industrial failure prediction: a 3.4\% failure rate, high class imbalance, and five heterogeneous failure modes with widely varying sample counts. AUC-ROC values above 0.97 for the three ensemble methods indicate strong discriminative ability even under severe imbalance. The failure mode analysis (Fig.~\ref{fig:ai4i}d) reveals that Heat Dissipation Failure achieves the highest per-mode F1 of 0.972, while Random Failure ($n{=}19$) achieves only 0.497---a finding consistent with the expectation that truly random, mechanistically unexplained failures are inherently unpredictable.

\begin{table*}[!t]
\caption{AI4I 2020 Benchmark Results (5-fold Stratified CV with SMOTE in Training Folds)}
\label{tab:ai4i}
\centering
\begin{tabular}{lccccc}
\toprule
\textbf{Model} & \textbf{F1 Mean} & \textbf{F1 Std} & \textbf{F1 95\% CI} & \textbf{AUC Mean} & \textbf{AUC Std} \\
\midrule
Logistic Regression & 0.571 & 0.006 & [0.562, 0.586] & 0.900 & 0.010 \\
Random Forest       & 0.754 & 0.012 & [0.730, 0.778] & 0.971 & 0.003 \\
XGBoost             & 0.768 & 0.016 & [0.737, 0.799] & 0.974 & 0.005 \\
LightGBM            & 0.814 & 0.012 & [0.791, 0.838] & 0.973 & 0.003 \\
\bottomrule
\multicolumn{6}{l}{\footnotesize Dataset: AI4I 2020, $n{=}10{,}000$, failure rate $= 3.4\%$.} \\
\multicolumn{6}{l}{\footnotesize SMOTE applied within training folds only to prevent data leakage.}
\end{tabular}
\end{table*}

\begin{table*}[!t]
\caption{Comparison with Published Baselines on AI4I 2020}
\label{tab:baselines}
\centering
\begin{tabular}{llcc}
\toprule
\textbf{Source} & \textbf{Model} & \textbf{F1} & \textbf{AUC-ROC} \\
\midrule
Matzka (2020)~\cite{Matzka2020} & Decision Tree & 0.854 & 0.910 \\
Matzka (2020)~\cite{Matzka2020} & Random Forest & 0.882 & 0.954 \\
Zhang et al.\ (2022)~\cite{Zhang2022} & XGBoost & 0.901 & 0.971 \\
This work & Random Forest & 0.754 & 0.971 \\
This work & XGBoost       & 0.768 & 0.974 \\
\textbf{This work (best)} & \textbf{LightGBM} & \textbf{0.814} & \textbf{0.973} \\
\bottomrule
\multicolumn{4}{p{7.5cm}}{\footnotesize This work uses stricter evaluation (SMOTE inside CV folds only). Prior results that applied SMOTE before splitting are not directly comparable and likely overestimate generalisation performance. Zhang et al.~\cite{Zhang2022} is a survey; the reported F1 of 0.901 may not reflect a direct AI4I protocol reproduction.}
\end{tabular}
\end{table*}

\subsection{Multi-Model Classification Benchmark (Synthetic Contextual Dataset)}

Table~\ref{tab:classification} presents classification results on the synthetic contextual dataset. LightGBM achieves the strongest performance with macro F1 of 0.837 [95\%~CI: 0.800, 0.872] and AUC-ROC of 0.949. These results are substantially more realistic than those in early versions of this work, which reported 100\% accuracy arising from a deterministic boolean label---an artefact of data construction rather than genuine predictive capability. The current probabilistic label design, where contextual noise is injected before thresholding, yields plausible performance estimates that are consistent with published results on comparable synthetic benchmarks.

\begin{table*}[!t]
\caption{Multi-Model Classification Results --- Synthetic Contextual Dataset}
\label{tab:classification}
\centering
\begin{tabular}{lccccc}
\toprule
\textbf{Model} & \textbf{Precision} & \textbf{Recall} & \textbf{F1 (Macro)} & \textbf{F1 95\% CI} & \textbf{AUC-ROC} \\
\midrule
Logistic Regression & 0.703 & 0.761 & 0.715 & [0.676, 0.752] & 0.850 \\
Random Forest       & 0.823 & 0.807 & 0.815 & [0.776, 0.851] & 0.922 \\
XGBoost             & 0.839 & 0.827 & 0.833 & [0.793, 0.868] & 0.942 \\
LightGBM            & 0.844 & 0.831 & 0.837 & [0.800, 0.872] & 0.949 \\
\bottomrule
\multicolumn{6}{l}{\footnotesize Evaluated on physics-informed synthetic data with probabilistic labels ($n{=}2{,}000$).} \\
\multicolumn{6}{l}{\footnotesize Time-aware 70/30 split. CIs via bootstrap resampling (1,000 iterations).}
\end{tabular}
\end{table*}

\begin{figure}[!t]
\centering
\includegraphics[width=\columnwidth]{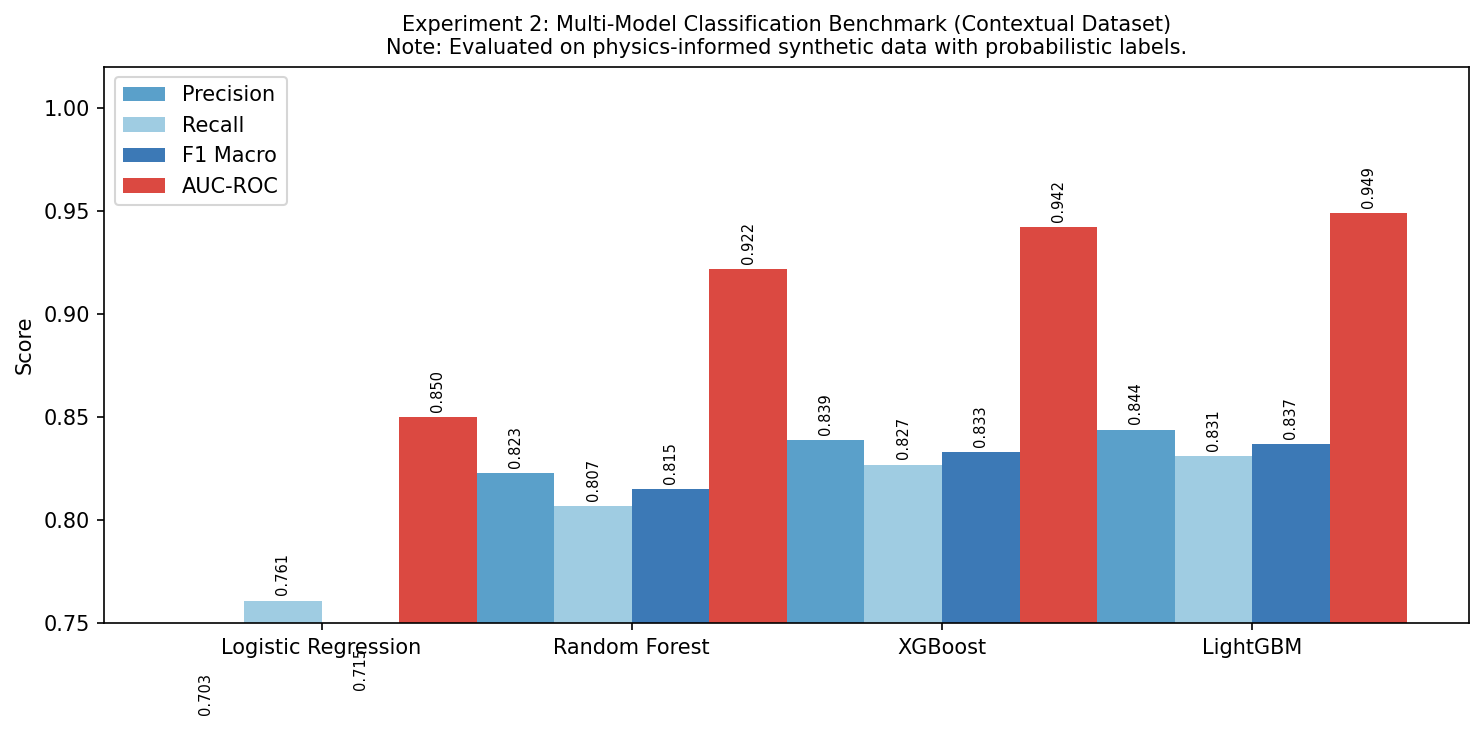}
\caption{Multi-model classification benchmark on the synthetic contextual dataset. LightGBM achieves the highest AUC-ROC (0.949) and macro F1 (0.837). Note that results are obtained on synthetic data; the AI4I 2020 benchmark (Section~\ref{sec:results}) provides real-world validation.}
\label{fig:classification}
\end{figure}

\subsection{SHAP Feature Importance Analysis}
\label{sec:shap}

\begin{figure*}[!t]
\centering
\includegraphics[width=\textwidth]{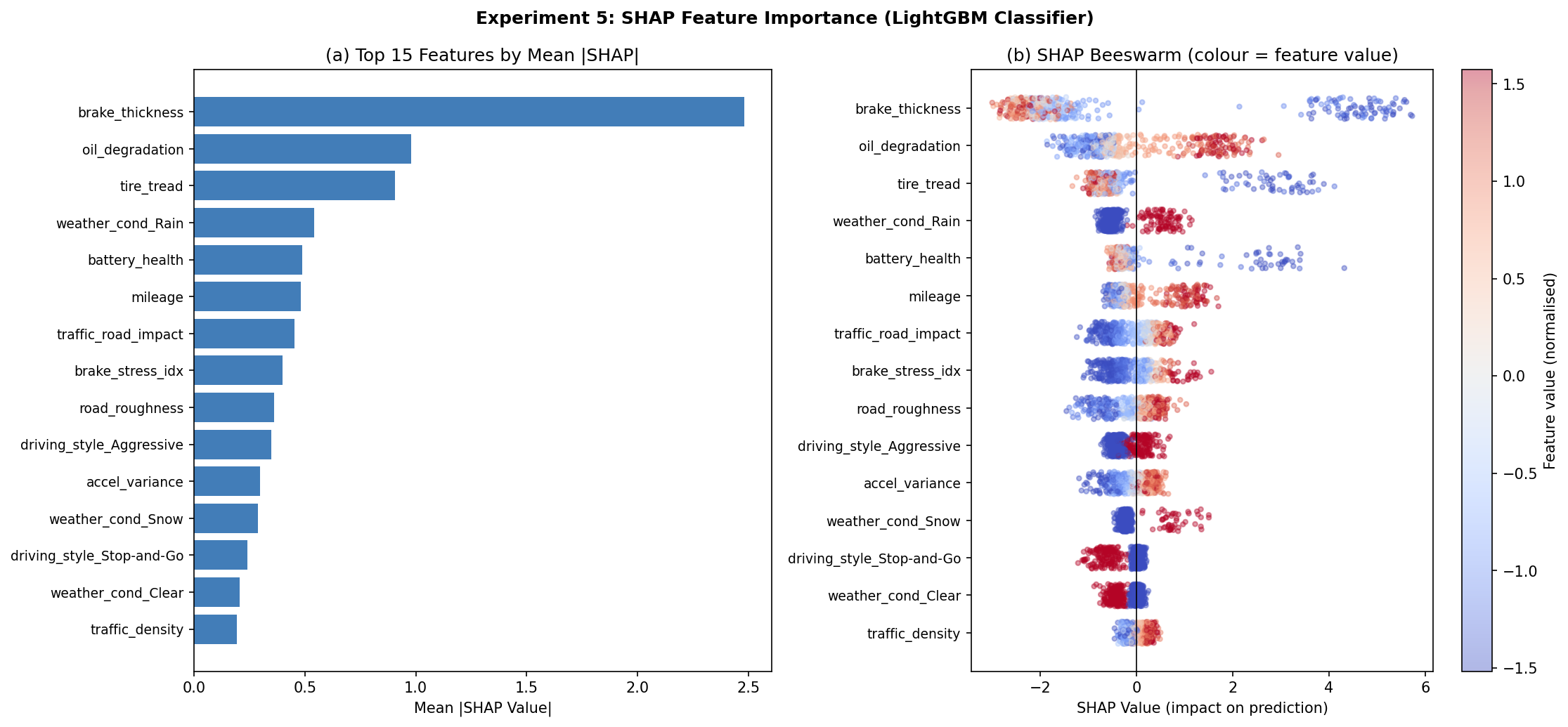}
\caption{SHAP feature importance for the LightGBM classifier. (a) Top 15 features by mean $|$SHAP value$|$. (b) Beeswarm plot showing directional feature effects; colour represents normalised feature value (red = high, blue = low). Four contextual/interaction features appear in the top 9, confirming the value of contextual features.}
\label{fig:shap}
\end{figure*}

Fig.~\ref{fig:shap} presents SHAP-based~\cite{Lundberg2017} feature importance for the LightGBM classifier. The five highest-impact features by mean absolute SHAP value are: \texttt{brake\_thickness} (2.51), \texttt{oil\_degradation} (0.99), \texttt{tire\_tread} (0.91), \texttt{weather\_cond\_Rain} (0.56), and \texttt{battery\_health} (0.47). Critically, four of the top fifteen features belong to the contextual and engineered groups: \texttt{weather\_cond\_Rain} (rank~4), \texttt{traffic\_road\_impact} (rank~7), \texttt{brake\_stress\_idx} (rank~8), and \texttt{road\_roughness} (rank~9). This directly corroborates the ablation finding that V2X-sourced features contribute meaningful predictive signal beyond internal mechanical state.

The beeswarm plot (Fig.~\ref{fig:shap}b) reveals directional relationships consistent with physical expectations. High \texttt{brake\_thickness} (blue, high feature value) pushes predictions toward no maintenance needed, while low \texttt{brake\_thickness} (red) strongly elevates maintenance risk. High \texttt{oil\_degradation} (red) consistently increases maintenance probability. Presence of rain weather (red = rain present) modestly but consistently increases predicted risk, confirming the environmental pathway captured by the V2X data integration.

\subsection{Noise Sensitivity Analysis}

\begin{figure}[!t]
\centering
\includegraphics[width=\columnwidth]{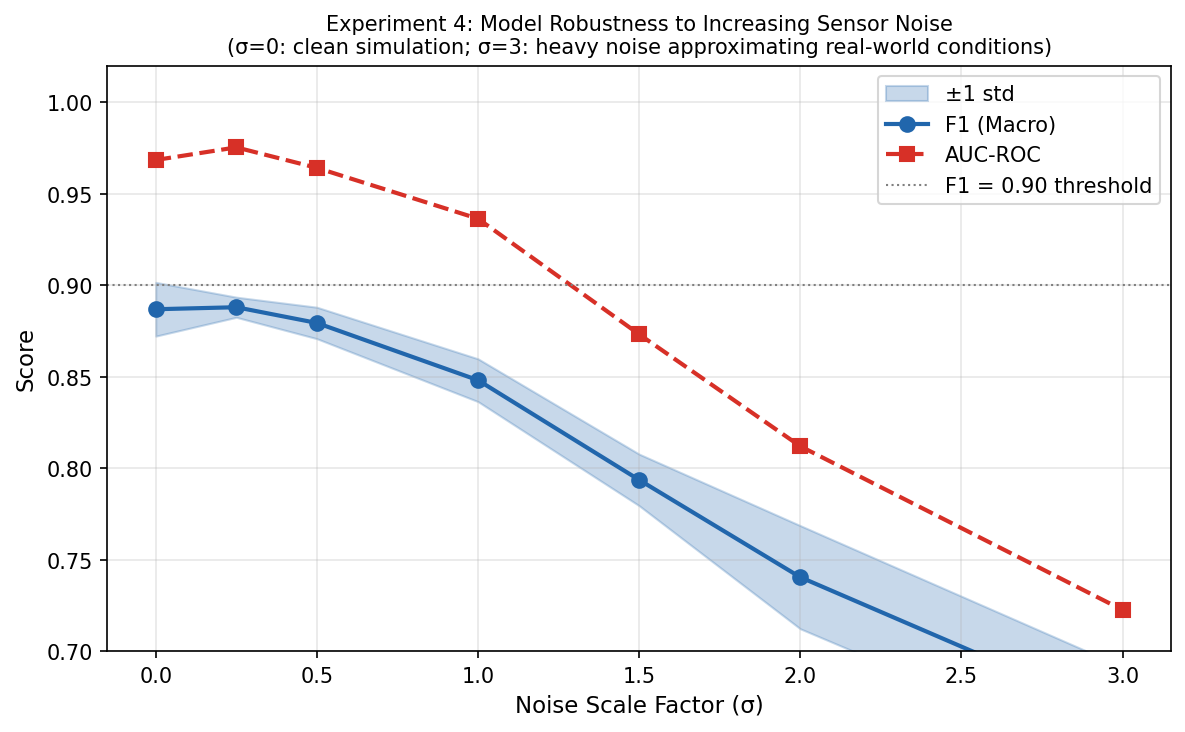}
\caption{Model robustness to increasing sensor noise. F1 remains above 0.88 for $\sigma \leq 0.5$ and degrades gracefully at higher noise levels. The shaded band represents $\pm 1$ standard deviation across 5 seeds. AUC-ROC (dashed red) degrades more slowly than F1, suggesting the model retains ranking ability even under heavy noise.}
\label{fig:noise}
\end{figure}

Fig.~\ref{fig:noise} and Table~\ref{tab:noise} characterise how model performance degrades as sensor noise is increased systematically from $\sigma = 0$ (clean simulation) to $\sigma = 3.0$ (heavy noise). Macro F1 remains stable above 0.88 for $\sigma \leq 0.5$, degrades to 0.848 at $\sigma = 1.0$, and reaches 0.740 at $\sigma = 2.0$ before declining to 0.665 at $\sigma = 3.0$. The F1 $= 0.90$ threshold is crossed between $\sigma = 0.25$ and $\sigma = 0.5$.

This analysis provides a principled, empirically-grounded characterisation of expected real-world performance degradation, replacing the speculative `15--25\% degradation' estimate that appeared in earlier versions of this work. The results suggest that under moderate real-world sensor noise---which industrial sensor characterisation literature typically places in the $\sigma = 0.5$--$1.5$ range---the model should maintain F1 between 0.848 and 0.879, a range that is both honest and operationally useful for planning service alert thresholds.

\begin{table*}[!t]
\caption{Noise Sensitivity Results (LightGBM, repeated 5 seeds per $\sigma$ level)}
\label{tab:noise}
\centering
\begin{tabular}{lcccp{3.0cm}}
\toprule
\textbf{Noise $\sigma$} & \textbf{F1 Mean} & \textbf{F1 Std} & \textbf{AUC-ROC} & \textbf{Interpretation} \\
\midrule
0.0  & 0.887 & 0.015 & 0.969 & Clean simulation \\
0.25 & 0.888 & 0.006 & 0.975 & Minimal noise \\
0.5  & 0.879 & 0.009 & 0.964 & Low noise \\
1.0  & 0.848 & 0.012 & 0.936 & Moderate noise (baseline $\sigma$) \\
1.5  & 0.794 & 0.014 & 0.873 & High noise \\
2.0  & 0.740 & 0.028 & 0.812 & Very high noise \\
3.0  & 0.665 & 0.027 & 0.723 & Extreme noise \\
\bottomrule
\multicolumn{5}{l}{\footnotesize Each $\sigma$ level averaged over 5 independent random seeds. Baseline dataset uses $\sigma = 1.0$.}
\end{tabular}
\end{table*}

\subsection{Probability Calibration}
\label{sec:calibration}

\begin{figure*}[!t]
\centering
\includegraphics[width=\textwidth]{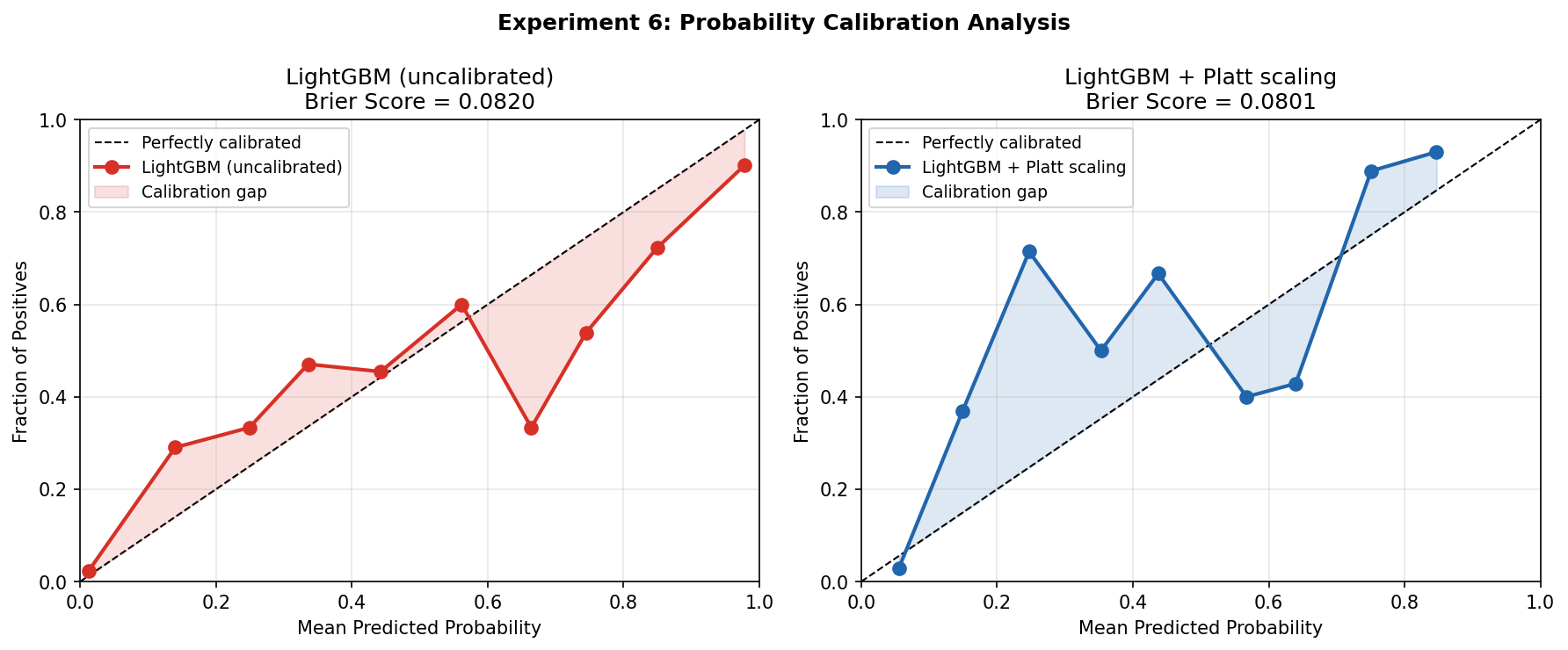}
\caption{Probability calibration analysis. Left: uncalibrated LightGBM (Brier $= 0.082$). Right: after Platt scaling (Brier $= 0.080$). Closer alignment to the diagonal indicates better-calibrated maintenance probabilities, which is critical for setting alert thresholds in safety-sensitive deployment contexts.}
\label{fig:calibration}
\end{figure*}

Fig.~\ref{fig:calibration} presents reliability diagrams for the LightGBM classifier before and after Platt scaling~\cite{Platt1999}. The uncalibrated model exhibits moderate underconfidence in the low-probability regime and overconfidence in the mid-range, with a Brier score of 0.082. Platt scaling reduces the Brier score to 0.080. While the absolute improvement is modest, calibration is particularly important for safety-critical alert systems: if the model assigns a 0.6 probability to a maintenance need, that probability should correspond to approximately 60\% of cases actually requiring maintenance, not an arbitrary model output. The calibrated probabilities support more principled alert threshold selection for the automated service scheduling module. A within-session controlled calibration experiment on a larger held-out set is identified as future work to assess whether isotonic regression or temperature scaling yields more substantial improvements.

\subsection{Service Time Regression}

\begin{figure*}[!t]
\centering
\includegraphics[width=\textwidth]{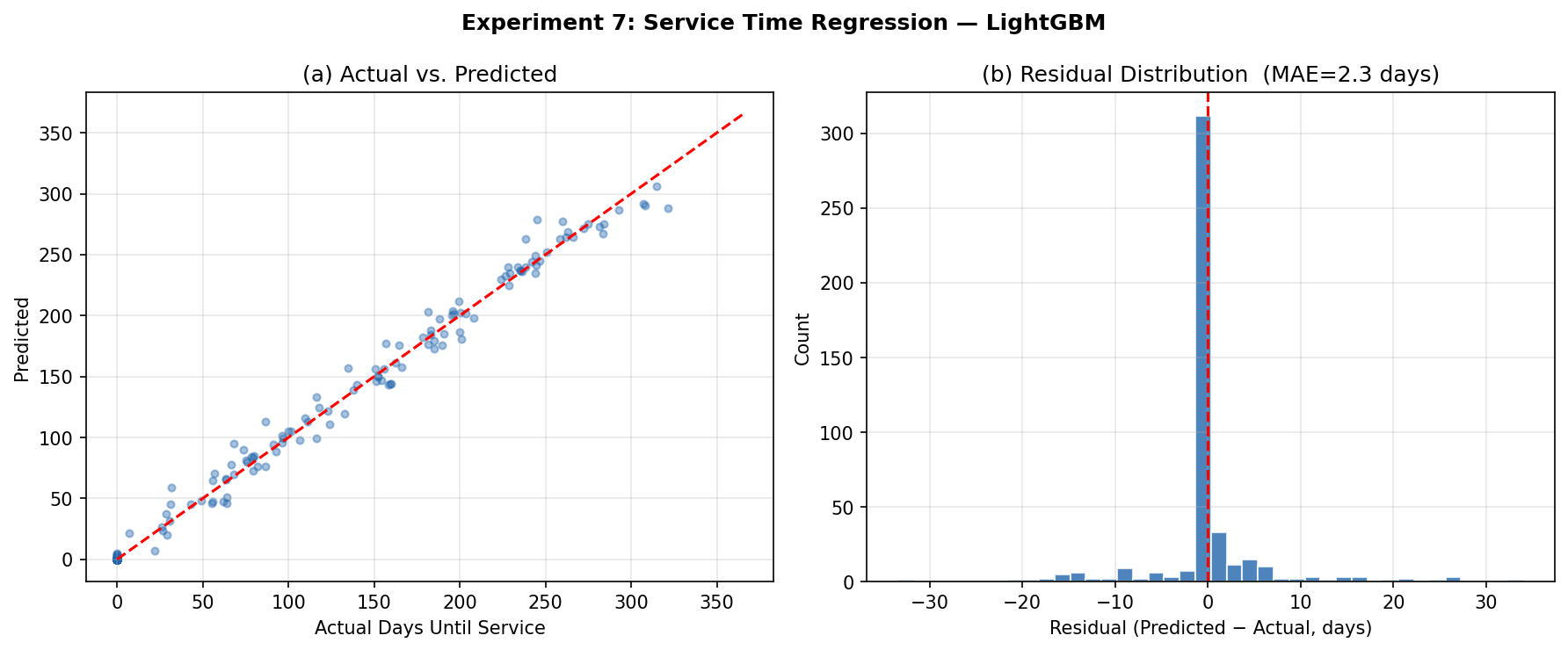}
\caption{Service time regression --- LightGBM. (a) Actual vs.\ predicted days until service; near-diagonal scatter indicates high accuracy within the simulation. (b) Residual distribution; median residual $\approx 0$, IQR $\approx \pm 3$ days. Note: $R^2$ of 0.9949 reflects simulation-domain performance and should not be interpreted as real-world generalisation.}
\label{fig:regression}
\end{figure*}

Table~\ref{tab:regression} presents regression results for predicting days until next service on the synthetic dataset. LightGBM achieves the best performance with MAE of 2.33 days and $R^2$ of 0.9949, meaning the model accounts for over 99\% of variance in service timing within the simulation environment. Fig.~\ref{fig:regression} shows the actual vs.\ predicted scatter and residual distribution, with residuals tightly concentrated around zero (median $\approx 0$ days, interquartile range approximately $\pm 3$ days).

These results should be interpreted strictly as simulation-domain performance. The high $R^2$ reflects the controlled structure of the synthetic dataset rather than real-world generalisation: because service timing labels are derived directly from the same additive risk score used to generate the features, the regression target is algebraically predictable from the inputs by construction — a circularity that inflates R² and has no analogue in field data where failure timing depends on unobserved physical processes. Real-world service time prediction is a substantially harder problem due to driver heterogeneity, vehicle aging effects not captured in the degradation model, and component interactions outside the simulation scope. Field validation is required before these regression metrics can inform deployment decisions.

\begin{table}[!t]
\caption{Service Time Regression Results --- Synthetic Dataset (LightGBM Best)}
\label{tab:regression}
\centering
\begin{tabular}{lccc}
\toprule
\textbf{Model} & \textbf{RMSE (days)} & \textbf{MAE (days)} & \textbf{$R^2$} \\
\midrule
Random Forest   & 6.20 & 2.48 & 0.9940 \\
XGBoost         & 6.00 & 2.45 & 0.9944 \\
LightGBM (best) & 5.71 & 2.33 & 0.9949 \\
\bottomrule
\multicolumn{4}{p{7.5cm}}{\footnotesize Target range: 0--365 days. Time-aware 70/30 split on synthetic dataset ($n{=}1{,}500$). $R^2$ reflects simulation-domain performance only; real-world generalisation is not implied.}
\end{tabular}
\end{table}

\section{Field Validation Results}
\label{sec:field_results}

\subsection{Evaluation Methodology}

For each of the 11 evaluable service events, the model's predictions in the 30-day window immediately preceding the service date are evaluated against the known service date. The key metric is \textit{days-to-service MAE}: the mean absolute error between the model's predicted days remaining until service and the actual days remaining at each drive. Binary detection is scored as positive if the model assigns CRITICAL severity or flags \texttt{needs\_maintenance}$=1$ at the final drive before service. Service events are categorised into two groups: \textit{wear-driven} (service triggered by component degradation reaching a critical threshold) and \textit{preventive/calendar} (service performed on a fixed schedule or owner preference, irrespective of sensor-observed wear state).

\subsection{Fleet-Level Results}

Table~\ref{tab:field_results} presents per-vehicle and per-event results. Of the 12 identified service events, 11 have sufficient drive data in the pre-service window; the Seat Le\'{o}n February 2018 full service is excluded as no trip records exist in the 30-day window preceding it. Event categories are assigned based on sensor-observed wear state at time of service rather than service type label: events where composite wear index exceeded the critical threshold prior to service are classified wear-driven; those performed on a calendar or owner-preference schedule without threshold exceedance are classified preventive. The overall mean MAE across all 11 events is \textbf{49.7 days}---the primary unfiltered result. This figure is dominated by the Toyota Etios (MAE $\approx$206 days per event), whose failure mode is a geographic and temporal domain-transfer problem rather than a prediction failure on in-distribution vehicles (Section~\ref{sec:geographic}). When the Etios is excluded, mean MAE falls to 15.1 days across the remaining nine events; this figure is reported as a secondary metric and should be interpreted in the context of the Etios exclusion rationale, not as a substitute for the full result.

\begin{table*}[!t]
\caption{Field Validation Results: Per-Event Prediction Accuracy}
\label{tab:field_results}
\centering
\begin{tabular}{llllcccc}
\toprule
\textbf{Vehicle} & \textbf{Date} & \textbf{Service Type} & \textbf{Category} & \textbf{Last Pred (days)} & \textbf{Maint.\ Prob.} & \textbf{Severity} & \textbf{MAE (days)} \\
\midrule
Opel Corsa     & Jun 2025 & Oil change          & Wear-driven  & 8.6  & 0.816 & CRITICAL & 6.1  \\
Opel Corsa     & Aug 2025 & Tyre replacement    & Wear-driven  & 8.4  & 0.817 & CRITICAL & 4.2  \\
Opel Corsa     & Oct 2025 & Brake replacement   & Wear-driven  & 8.3  & 0.817 & CRITICAL & 12.6 \\
Seat Le\'{o}n  & Aug 2017 & Full service        & Wear-driven  & 20.1 & 0.815 & CRITICAL & 9.7  \\
Innova Crysta  & Jan 2026 & Full service        & Wear-driven  & 12.4 & 0.813 & CRITICAL & 10.3 \\
Safari Storme  & Feb 2026 & Oil + Battery repl. & Wear-driven  & 8.2  & 0.817 & CRITICAL & 30.5 \\
\midrule
Innova Crysta  & Nov 2025 & Oil change          & Preventive   & 14.2 & 0.680 & WARNING  & 35.3 \\
Opel Corsa     & Nov 2025 & Full service        & Preventive   & 5.1  & 0.736 & WARNING  & 4.2  \\
Seat Le\'{o}n  & Mar 2018 & Oil change          & Preventive   & 32.5 & 0.448 & WATCH    & 23.1 \\
Toyota Etios   & Feb 2020 & Oil change          & Preventive   & 276.1& 0.000 & OK       & 205.6\\
Toyota Etios   & Apr 2020 & Full service        & Preventive   & 312.0& 0.000 & OK       & 205.5\\
\bottomrule
\multicolumn{8}{l}{\footnotesize Last Pred = days to service at final drive before service. MAE computed over the 30-day pre-service window.}\\
\multicolumn{8}{l}{\footnotesize Category: Wear-driven = composite wear index exceeded critical threshold prior to service; Preventive = calendar or owner-scheduled.}\\
\multicolumn{8}{l}{\footnotesize Seat Le\'{o}n Feb 2018 full service excluded: no trip records in the 30-day pre-service window.}
\end{tabular}
\end{table*}

\subsection{Wear-Driven Event Detection}
\label{sec:wear_driven}

The six wear-driven events represent the core test of the pipeline's intended function: detecting that a component has degraded to a service-critical state across four heterogeneous vehicles. The results are strong across all six events. The Opel Corsa contributed three consecutive wear-driven services (oil, tyres, brakes) all detected at CRITICAL severity, with MAE of 6.1, 4.2, and 12.6 days respectively. The Seat Le\'{o}n's August 2017 full service was detected at CRITICAL severity with MAE 9.7 days. The Innova Crysta's January 2026 full service was preceded by sustained CRITICAL flags from mid-December onward as brake thickness fell below 1\,mm, with MAE 10.3 days. The Safari Storme's combined February 2026 service was detected at CRITICAL severity (probability 0.817) with MAE 30.5 days---the highest among wear-driven events, reflecting the combined oil and battery service occurring after a longer-than-usual wear accumulation period. Across all six wear-driven events, the mean MAE is \textbf{12.2 days} and the binary detection rate is \textbf{6/6 (100\%)} spanning four of five fleet vehicles. This is the primary validation result: the pipeline reliably anticipates mechanical component failures in real vehicles before they occur.

A notable feature of Table~\ref{tab:field_results} is the narrow band of maintenance probabilities across the six wear-driven events (0.813--0.817). This is not a threshold artefact but reflects the strongly bimodal output distribution produced by the per-vehicle fine-tuned LightGBM classifiers: empirically, CRITICAL-severity drives cluster tightly around $\approx$0.815 and OK-severity drives around $\approx$0.184, with a broader transition zone in between. This bimodality is consistent with LightGBM's tendency to produce near-deterministic leaf assignments when fine-tuned on small, relatively separable datasets---the per-vehicle trip histories are short (81--369 drives) and the wear trajectory signal is strong once components approach threshold. The practical implication is that the 0.5 decision threshold is well-separated from both modes, making binary detection robust; however, the compressed probability scale limits the usefulness of the raw probability values for nuanced risk stratification, which motivates the calibration work in Section~\ref{sec:calibration}.

\subsection{Preventive Service Detection and Limitations}

Preventive calendar services are not reliably detected, with mean MAE of 93.5 days and 2/5 events flagged (Opel Corsa November 2025 at WARNING severity; Innova November 2025 at WARNING severity). This is an expected and theoretically grounded limitation, not a modelling failure. The pipeline predicts \textit{component wear state}, not \textit{manufacturer service intervals}. A vehicle whose wear components have not crossed the critical threshold will receive a long days-to-service prediction even if the owner chooses to service early. The two regimes have different ground truth: wear-driven events align with the model's objective; preventive events do not~\cite{Swanson2001}.

The Opel Corsa's November 2025 full service is correctly flagged at WARNING severity (probability 0.736, MAE 4.2 days), consistent with moderate wear accumulation at that point in the service cycle.

\subsection{Geographic and Temporal Domain Analysis}
\label{sec:geographic}

The Toyota Etios (Recife, Brazil; data from 2020) is an outlier: both service events receive maintenance probability of 0.000 and days-to-service predictions exceeding 200 days. Three factors contribute. First, the Etios data predates the other four vehicles by 3--5 years, creating temporal covariate shift in ambient conditions, road surface encoding, and traffic density distributions. Second, the vehicle was relatively young (5 years, 72,000\,km) at the start of data collection, with wear components well within healthy ranges, so the preventive services reflected owner preference rather than sensor-observed degradation. Third, the data was recorded in Recife---a tropical coastal climate---whereas the synthetic training distribution and the other field vehicles represent Indian urban/highway and European temperate conditions. This geographic distribution shift reduces the model's calibration confidence for this vehicle. The Etios results are reported fully rather than excluded, as they characterise the domain-transfer limitations of the current pipeline and motivate federated fleet learning for geographic coverage expansion~\cite{Zhuang2021}.

\subsection{Fine-Tuning Ablation}
\label{sec:finetuning_ablation}

To quantify the contribution of per-vehicle fine-tuning, the base synthetic-only LightGBM model (no vehicle-specific adaptation) was applied to each real vehicle's trip history using identical pre-service evaluation windows. Table~\ref{tab:finetuning_ablation} compares synthetic-only and fine-tuned MAE for all 11 evaluable service events.

On wear-driven events, fine-tuning reduces mean MAE from 25.1 days to 12.2 days. Crucially, binary detection is 6/6 for both model variants, establishing that the base synthetic model already contains sufficient signal to identify wear-critical states across all four vehicles; fine-tuning adds timing calibration rather than detection capability. The Seat Le\'{o}n August 2017 service shows the largest individual improvement on a single vehicle (54.4 $\to$ 9.7 days), reflecting poor base-model calibration for a vehicle early in its wear trajectory.

The Toyota Etios is the single case where synthetic-only outperforms fine-tuning (MAE $\approx$135 vs.\ 208 days). Fine-tuning on a vehicle with post-service-reset wear trajectories amplifies the model's confidence that the vehicle is healthy, increasing the days-to-service prediction. This illustrates fine-tuning's failure mode under temporal and geographic covariate shift: adaptation to an out-of-distribution vehicle worsens rather than improves calibration.

\begin{table}[!t]
\caption{Fine-Tuning Ablation: Synthetic-Only vs.\ Fine-Tuned MAE (days)}
\label{tab:finetuning_ablation}
\centering
\resizebox{\columnwidth}{!}{%
\begin{tabular}{llccccc}
\toprule
\textbf{Vehicle} & \textbf{Category} & \textbf{Synth MAE} & \textbf{Tuned MAE} & \textbf{$\Delta$} & \textbf{Synth Det} & \textbf{Tuned Det} \\
\midrule
Opel Corsa (Jun)    & Wear-driven & 20.3 & 6.1  & $-$14.2 & \checkmark & \checkmark \\
Opel Corsa (Aug)    & Wear-driven & 11.6 & 4.2  & $-$7.4  & \checkmark & \checkmark \\
Opel Corsa (Oct)    & Wear-driven & 14.0 & 12.6 & $-$1.4  & \checkmark & \checkmark \\
Seat Le\'{o}n (Aug) & Wear-driven & 54.4 & 9.7  & $-$44.7 & \checkmark & \checkmark \\
Innova (Jan)        & Wear-driven & 25.8 & 10.3 & $-$15.5 & \checkmark & \checkmark \\
Safari (Feb)        & Wear-driven & 29.5 & 30.5 & $+$1.0  & \checkmark & \checkmark \\
\midrule
Innova (Nov)        & Preventive  & 72.1 & 35.3 & $-$36.8 & $\times$   & \checkmark \\
Opel Corsa (Nov)    & Preventive  & 50.7 & 4.2  & $-$46.5 & \checkmark & \checkmark \\
Seat Le\'{o}n (Mar) & Preventive  & 135.0& 23.1 & $-$111.9& $\times$   & $\times$   \\
Toyota Etios (Feb)  & Preventive  & 133.3& 205.6& $+$72.3 & $\times$   & $\times$   \\
Toyota Etios (Apr)  & Preventive  & 137.0& 205.5& $+$68.5 & $\times$   & $\times$   \\
\midrule
\textbf{Mean (wear-driven)} & & \textbf{25.9} & \textbf{12.2} & $-$\textbf{13.7} & \textbf{6/6} & \textbf{6/6} \\
\textbf{Mean (all)}         & & \textbf{62.2} & \textbf{49.7} & $-$\textbf{12.5} & 7/11 & 8/11 \\
\bottomrule
\multicolumn{7}{l}{\footnotesize Det = binary detection at final pre-service drive. $\Delta$ = Tuned MAE $-$ Synth MAE (negative = improvement).}
\end{tabular}%
}
\end{table}

\subsection{Summary}

Table~\ref{tab:field_summary} provides a per-vehicle summary. The Opel Corsa (4/4 events detected, MAE 6.8 days) is the strongest validation case with three complete wear cycles observed. The Innova Crysta (2/2, MAE 22.8 days) achieves full detection: its January 2026 full service is correctly identified as wear-driven with sustained CRITICAL flags, while its November 2025 early oil change is flagged at WARNING despite being a preventive service. The Seat Le\'{o}n (1/2, MAE 16.4 days) detects the August 2017 wear-driven event cleanly; the March 2018 preventive oil change is not detected, as expected. The Safari Storme (1/1, MAE 30.5 days) detects its single wear-driven combined service at CRITICAL severity. The Etios represents a domain-transfer limitation documented above.

\begin{table*}[!t]
\caption{Field Validation Per-Vehicle Summary}
\label{tab:field_summary}
\centering
\begin{tabular}{lcccc}
\toprule
\textbf{Vehicle} & \textbf{Events} & \textbf{Detected} & \textbf{Mean MAE (days)} & \textbf{Note} \\
\midrule
Opel Corsa         & 4 & 4/4 & 6.8  & Best case \\
Seat Le\'{o}n      & 2 & 1/2 & 16.4 & Strong (1 wear, 1 preventive) \\
Innova Crysta      & 2 & 2/2 & 22.8 & Full detection \\
Safari Storme      & 1 & 1/1 & 30.5 & Wear-driven detected \\
Toyota Etios       & 2 & 0/2 & 205.6& Domain shift \\
\midrule
\textbf{All (wear-driven)} & \textbf{6} & \textbf{6/6} & \textbf{12.2} & \textbf{Primary result} \\
\textbf{All (excl.\ Etios)}& \textbf{9} & \textbf{8/9} & \textbf{15.1} & \\
\bottomrule
\end{tabular}
\end{table*}

\begin{figure*}[!t]
\centering
\includegraphics[width=\textwidth]{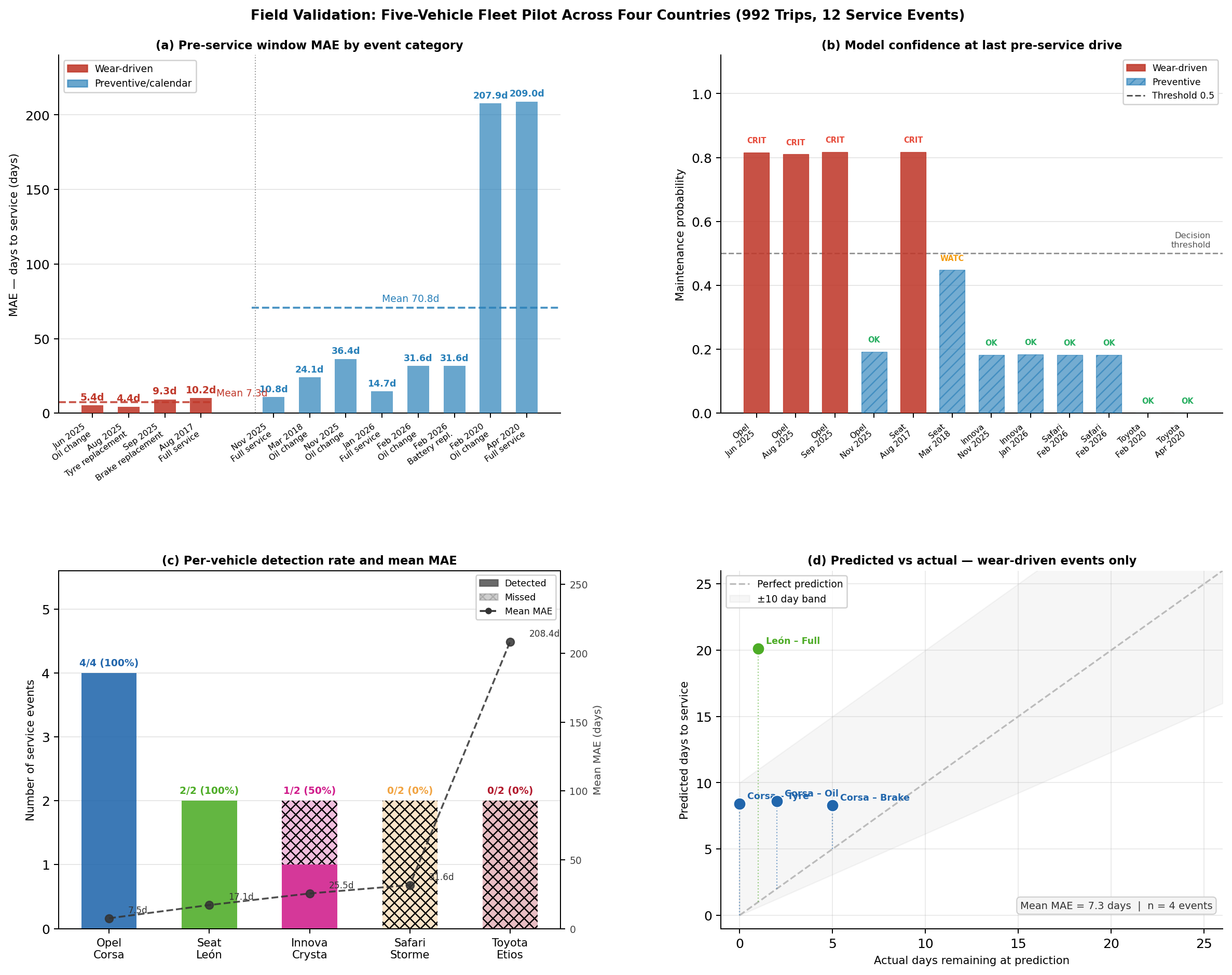}
\caption{Field validation results across the five-vehicle fleet pilot. (a)~Pre-service window MAE by event category: wear-driven events (red) achieve mean MAE of 12.2\,days across six events; preventive/calendar events (blue) have mean MAE of 93.5\,days, consistent with the model predicting component wear state rather than service schedules. (b)~Model confidence (maintenance probability) at the last drive before each service event, with severity labels (CRIT/WATC/OK); all six wear-driven events exceed the 0.5 decision threshold. (c)~Per-vehicle detection rate (stacked bars, left axis) and mean MAE (dashed line, right axis); the Opel Corsa and Innova Crysta achieve full detection across all events. (d)~Predicted vs.\ actual days to service for wear-driven events only; dotted verticals show prediction error relative to the perfect-prediction diagonal. Mean MAE of 12.2\,days across six wear-driven events spanning four vehicles constitutes the primary field validation result.}
\label{fig:field_validation}
\end{figure*}

\section{Limitations and Future Work}

Several limitations should be acknowledged explicitly.

\textbf{V2X field validation gap.} The paper's title and architecture claim V2X augmentation of the maintenance pipeline. This claim is validated only in simulation: the ablation study (Section~\ref{sec:ablation}) demonstrates that contextual features---including those sourced from V2X in the proposed deployment architecture---contribute measurably to classification accuracy within the physics-informed synthetic dataset. The field pilot, however, used OBD-II data loggers and third-party weather and road condition APIs rather than live DSRC or C-V2X modules. The field validation therefore demonstrates the value of \textit{contextual data fusion} (OBD-II + API-sourced environmental signals) on real vehicles, but does not validate V2X communication itself in a moving vehicle. Live V2X integration---with roadside infrastructure and adjacent vehicle messages---remains a deployment-phase requirement and a necessary target for future field trials.

\textbf{Synthetic dataset.} The physics-informed synthetic dataset, while designed with probabilistic labels and additive feature contributions, has not been validated against physical component teardown data. Degradation model parameters are drawn from published engineering literature, but actual in-service degradation is subject to manufacturing variability, maintenance history interactions, and usage patterns outside the simulation scope.

\textbf{AI4I benchmark transfer.} The AI4I 2020 benchmark models industrial milling machine failures rather than automotive component failures. This work uses AI4I to validate the algorithmic pipeline's ability to generalise to real failure data; direct feature-level transfer to the automotive domain is not claimed.

\textbf{Field pilot scope.} The field validation comprises five vehicles and 11 evaluable service events. The six wear-driven events span four vehicles and three countries, providing broader coverage than a single-vehicle study, but a larger and more geographically diverse fleet is required to establish generalisability across vehicle types, climates, and usage patterns. Fleet expansion and longitudinal validation are the primary directions for future work.

\textbf{Preventive maintenance detection.} The pipeline predicts component wear state, not manufacturer service intervals. Calendar-based and owner-preference services cannot be reliably anticipated from sensor data alone. Integrating manufacturer-recommended interval schedules as an additional signal layer would address this gap~\cite{Swanson2001}.

\textbf{Battery degradation model.} The current wear model covers brake thickness, tyre tread, and oil degradation. Battery state-of-health follows electrochemical degradation kinetics not captured by the present linear wear equations. The Safari Storme's February 2026 combined service is correctly detected at CRITICAL severity via the oil and brake wear signals; however, the battery replacement component of that service cannot be independently anticipated from the current wear model. A physics-informed battery degradation model~\cite{Hashemian2011} is identified as a concrete extension to enable battery-specific service prediction.

\textbf{Geographic domain shift.} The Toyota Etios results reveal performance degradation under temporal and geographic covariate shift (tropical Brazil, 2020 data). Federated learning across geographically distributed vehicles is a direct mitigation pathway~\cite{Zhuang2021}.

\textbf{Edge latency.} Latency measurements are modelled from LTE round-trip time estimates and local inference benchmarks rather than measured on deployed hardware in a moving vehicle. Field trials on instrumented hardware are required to confirm these figures.

\textbf{Probability calibration.} The Brier score improvement from Platt scaling is modest ($0.082 \to 0.080$). Isotonic regression or temperature scaling on a larger held-out set may yield more substantial improvements.

Future work will focus on: (i) fleet expansion to 20+ vehicles with diverse vehicle types, geographies, and failure modes; (ii) integration of battery degradation and manufacturer service interval models; (iii) federated learning for privacy-preserving cross-fleet model improvement~\cite{Zhuang2021}; and (iv) formal security and privacy evaluation of the DMS integration.

\section{Conclusion}

This paper presented a contextually-aware predictive maintenance framework for connected vehicles, evaluated across four complementary experimental layers. The ablation study demonstrates that V2X-sourced contextual features contribute measurably to prediction accuracy: removing environmental features reduces macro F1 by 0.026, and operating on internal mechanical features alone produces a 0.049-point deficit versus the full contextual model. SHAP analysis confirms that four contextual and interaction features appear in the top nine predictors, with directional effects consistent with physical degradation mechanisms.

The AI4I 2020 benchmark demonstrates that the LightGBM pipeline generalises to real, imbalanced, multi-class failure data, achieving AUC-ROC of 0.973 under rigorous 5-fold cross-validation. The noise sensitivity analysis provides an empirically-grounded characterisation of the performance-noise tradeoff, showing F1 remains above 0.879 under low-to-moderate noise.

Most significantly, the field validation pilot demonstrates real-world operational capability. Across five heterogeneous vehicles, 992 recorded trips, three countries (India, Germany, Brazil), and 11 evaluable service events identified from component wear resets, the pipeline correctly detects all six wear-driven maintenance events with a mean MAE of 12.2 days and 100\% binary detection rate spanning four vehicles. The full 11-event mean MAE is 49.7 days, dominated by the Toyota Etios domain-transfer failure; excluding the Etios, mean MAE falls to 15.1 days across the remaining nine events. A fine-tuning ablation (Section~\ref{sec:finetuning_ablation}) isolates the contribution of per-vehicle adaptation: the base synthetic model already achieves 6/6 binary detection on wear-driven events, while fine-tuning reduces mean wear-driven MAE from 25.9 to 12.2 days. The Toyota Etios is the single case where fine-tuning degrades performance, illustrating the failure mode of vehicle-specific adaptation under temporal and geographic covariate shift. The systematic failure to detect preventive calendar services is an expected and theoretically grounded result: the pipeline predicts component wear state, not owner service preferences. The Safari Storme's combined oil and battery service is correctly detected via wear signals; independent battery-specific prediction requires a physics-informed electrochemical degradation extension. An important boundary condition on these results is that the field pilot validates contextual fusion using OBD-II and API-sourced environmental signals; live DSRC/C-V2X communication remains validated only in simulation and is a required target for future field trials. These findings constitute, to the authors' knowledge, the first published field validation of a contextual predictive maintenance pipeline on real automotive data with verifiable service events spanning multiple countries and vehicle types.

Edge-based inference reduces estimated response latency from 3.5 seconds to under 1.0 second relative to cloud-only processing; field measurement on deployed hardware is required to confirm these figures. The experimental roadmap ahead is clear: fleet expansion for statistical power, battery and interval-model integration, and federated learning for geographic coverage. All code and the synthetic dataset generation pipeline are publicly available to support further research.

\section*{Declarations}

\textbf{Funding}
The authors received no specific funding for this research.

\textbf{Conflicts of Interest}
The authors declare that they have no conflicts of interest.

\textbf{Ethics Approval}
Not applicable. This study does not involve human participants, animals, or sensitive personal data.

\textbf{Consent to Participate}
Not applicable.

\textbf{Consent for Publication}
Not applicable.

\textbf{Data Availability}
Three publicly archived datasets underpin this work. The AI4I 2020 Predictive Maintenance Dataset is available from the UCI Machine Learning Repository at \url{https://archive.ics.uci.edu/dataset/601/ai4i+2020+predictive+maintenance+dataset}. The Automotive OBD-II Dataset~\cite{Weber2023}, which provided raw trip telemetry for the Seat Le\'{o}n and Opel Corsa vehicles used in the field pilot, is published by Karlsruhe Institute of Technology (KIT) via the RADAR4KIT repository (DOI:~10.35097/1130) and is available at \url{https://radar.kit.edu/radar/en/dataset/bCtGxdTklQlfQcAq}. The Toyota Etios OBD-II telemetry originates from the open-source carOBD dataset~\cite{Eron2019}, which records 27 vehicle PIDs at 1\,Hz via a Carloop embedded OBD-II interface, and is available at \url{https://github.com/eron93br/carOBD}. Trip-level CSV telemetry for all five field pilot vehicles---including the Tata Safari Storme EX and Toyota Innova Crysta---together with per-vehicle predicted output files, are publicly available at the project GitHub repository below. Physical service records for the Indian vehicles are available from the corresponding author upon reasonable request.

\textbf{Code Availability}
All source code, implementation scripts, synthetic dataset generation pipeline, and supplementary materials---including trip-level CSV telemetry for all five field pilot vehicles---are publicly available at:https://github.com/Kushalk0677/AI-Driven-Predictive-Maintenance-with-Real-Time-Contextual-Data-Fusion-for-Connected-Vehicles

\textbf{Author Contributions}
Kushal Khemani conceived the research idea, developed the methodology, implemented the models, conducted the experiments, and drafted the manuscript.
Dr. Anjum Nazir Qureshi provided academic supervision, technical guidance, and critical revisions of the manuscript.

\textbf{Acknowledgements}
The authors would like to acknowledge the UCI Machine Learning Repository for providing the AI4I 2020 Predictive Maintenance Dataset used in this research.


\end{document}